\pdfoutput=1
\documentclass[11pt,a4paper]{article}
\usepackage{authblk}
\usepackage[hyperref]{acl2020}
\usepackage{times}
\usepackage{latexsym}

\usepackage{microtype}
\usepackage{breakurl}

\aclfinalcopy % Uncomment this line for the final submission
\def\aclpaperid{2476} %  Enter the acl Paper ID here

\usepackage{booktabs}
\usepackage{graphicx}
\usepackage[T1]{fontenc}
\newcommand*\rot{\rotatebox{90}}

\title{KLEJ: Comprehensive Benchmark for Polish Language Understanding}

\author[1]{Piotr Rybak}
\author[1]{Robert Mroczkowski}
\author[1]{Janusz Tracz}
\author[1,2]{Ireneusz Gawlik}
\affil[1]{
    ML Research at Allegro.pl, \protect\\
    ul. Grunwaldzka 182, 60\--166 Pozna\'{n}, Poland \protect\\
    \texttt{\{firstname.lastname\}@allegro.pl}\protect\\}
\affil[2]{
    AGH University of Science and Technology \protect\\
    Faculty of Computer Science, Electronics and Telecommunications, \protect\\
    Department of Computer Science \protect\\
    al. A. Mickiewicza 30, 30\--059 Krak\'{o}w, Poland}

\date{}

\begin{document}
\maketitle
\begin{abstract}
In recent years, a~series of Transformer-based models unlocked major improvements in general natural language understanding (NLU) tasks. Such a~fast pace of research would not be possible without general NLU benchmarks, which allow for a~fair comparison of the proposed methods. However, such benchmarks are available only for a~handful of languages. To alleviate this issue, we introduce a~comprehensive multi-task benchmark for the Polish language understanding, accompanied by an online leaderboard. It consists of a~diverse set of tasks, adopted from existing datasets for named entity recognition, question-answering, textual entailment, and others. We also introduce a~new sentiment analysis task for the e-commerce domain, named Allegro Reviews (AR). To ensure a~common evaluation scheme and promote models that generalize to different NLU tasks, the benchmark includes datasets from varying domains and applications. Additionally, we release HerBERT, a~Transformer-based model trained specifically for the Polish language, which has the best average performance and obtains the best results for three out of nine tasks. Finally, we provide an extensive evaluation, including several standard baselines and recently proposed, multilingual Transformer-based models.
\end{abstract}

\section{Introduction}

The field of natural language understanding (NLU) experienced a~major shift towards knowledge re-usability and transfer learning, a~phenomenon well established in the field of computer vision. Such a~shift was enabled by recent introduction of robust, general-purpose models suitable for fine-tuning, like ELMo \citep{Peters:2018}, ULMFiT \citep{howard2018universal} and BERT \citep{devlin2019bert}. These models significantly improved the state-of-the-art on numerous language understanding tasks. Since then, the progress accelerated significantly and the new Transformer-based \citep{vaswani2017attention} models are being published every month to claim the latest state-of-the-art performance.

Such a~pace of development would not be possible without standardized and publicly available NLU evaluation benchmarks. Among the most popular ones is the recently introduced GLUE \citep{wang2019glue} consisting of a~collection of tasks such as question answering, sentiment analysis, and textual entailment with texts coming from a~diverse set of domains. Some tasks come with numerous training examples, while others have limited training data. On top of that, for some tasks, the training set represents a~different domain than the test set. This promotes models that learn general language representations and are effective at transferring knowledge across various tasks and domains. The GLUE benchmark is constructed based on existing datasets, and its main contribution is the careful choice of tasks together with an online evaluation platform and a~leaderboard.

Unfortunately, most of the progress in NLU is happening for English and Chinese. Other languages lack both pretrained models and evaluation benchmarks. In this paper, we introduce the comprehensive multi-task benchmark for the Polish language understanding - KLEJ (eng. \emph{GLUE}, also abbreviation for \emph{Kompleksowa Lista Ewaluacji J\k{e}zykowych}, eng. \emph{Comprehensive List of Language Evaluations}). KLEJ consists of nine tasks and, similarly to GLUE, is constructed mostly out of existing datasets. The tasks were carefully selected to cover a~wide range of genres and different aspects of language understanding. Following GLUE, to simplify a~model evaluation procedure, we adjusted the tasks to fit into a~unified scoring scheme (either text classification or regression). Alongside the benchmark, we introduce HerBERT, a~Transformer-based model trained on several Polish text corpora. We compare HerBERT with a~set of both standard and recently introduced NLU baselines.

To summarize, our contributions are: 
\begin{enumerate}
    \item KLEJ: A~set of nine tasks constructed from both existing and newly introduced datasets used for the Polish language understanding evaluation,
    \item An online platform\footnote{\url{https://klejbenchmark.com}} to evaluate and present the model results in the form of a~leaderboard,
    \item HerBERT: Transformer-based model for the Polish language understanding,
    \item Allegro Reviews: A~new sentiment analysis task for the e-commerce domain,
    \item Evaluation of several LSTM-based baselines, multilingual Transformer-based models and HerBERT.
\end{enumerate}

The rest of the paper is organized as follows. In Section \ref{related}, we provide an overview of related work. In Section \ref{tasks}, we describe the tasks that make up the KLEJ benchmark. In Section \ref{baselines}, we give an overview of the selected baseline methods and introduce the new Transformer-based model for Polish. In Section \ref{evaluation}, we evaluate all models using KLEJ benchmark. Finally, we conclude our work in Section \ref{conclusion}.

\section{Related Work}
\label{related}

\begin{table*}
\centering
\setlength{\tabcolsep}{0.46em}
\renewcommand*{\arraystretch}{1.15}
\begin{tabular}{lrrrlll}

    \toprule
    \bf{Name} & \bf{Train} & \bf{Dev} & \bf{Test} & \bf{Domain} & \bf{Metrics} & \bf{Objective} \\
    \toprule
    \multicolumn{7}{c}{Single-Sentence Tasks} \\
    \midrule
    NKJP-NER & 16k & 2k & 2k & Balanced corpus & Accuracy & NER classification \\
    CDSC-R & 8k & 1k & 1k & Image captions & Spearman corr. & Semantic relatedness \\
    CDSC-E & 8k & 1k & 1k & Image captions & Accuracy & Textual entailment \\
    \midrule
    \multicolumn{7}{c}{Multi-Sentence Tasks} \\
    \midrule
    CBD & 10k & - & 1k & Social Media & F1-Score & Cyberbullying detection \\
    PolEmo2.0-IN & 6k & 0.7k & 0.7k & Online reviews & Accuracy & Sentiment analysis \\
    PolEmo2.0-OUT & 6k & 0.5k & 0.5k & Online reviews & Accuracy & Sentiment analysis \\
    Czy wiesz? & 5k & - & 1k & Wikipedia & F1-Score & Question answering \\
    PSC & 4k & - & 1k & News articles & F1-Score & Paraphrase \\
    AR & 10k & 1k & 1k & Online reviews & 1 $-$ wMAE & Sentiment analysis \\
    \bottomrule

\end{tabular}
\caption{The overview of tasks in the KLEJ benchmark. It consists almost exclusively of classification tasks, except for CDSC-R and AR which are regression.}
\label{tab:overview}
\end{table*}

The evaluation of NLU models was always an integral part of their development. Even though there are many established tasks on which to evaluate newly proposed models, there is no strict standard specifying which one to choose. The difficulty of a fair comparison between models eventually led to the introduction of multi-task benchmarks that unify the evaluation.

One such benchmark is SentEval \citep{conneau2018senteval}. It consists of seventeen established tasks used to evaluate the quality of sentence embeddings. Additionally, ten probing tasks are provided to detect what linguistic properties are retained in sentence embeddings. In all tasks, models take either a~single sentence embedding or a~pair of sentence embeddings as the input and solve a~classification (or a~regression) problem. The authors released a~toolkit\footnote{\url{https://github.com/facebookresearch/SentEval}} for model evaluation. However, they do not provide a~public leaderboard to compare the results of different models.

Another benchmark for evaluating models is decaNLP \citep{McCann2018decaNLP}, which consists of ten pre-existing tasks. In contrast to SentEval, choice of tasks is much more diverse, ranging from machine translation, semantic parsing to summarization. All tasks have been automatically converted to a~question answering format. 

Finally, the GLUE benchmark \citep{wang2019glue} proposes a~set of nine tasks. All of them are constructed from existing, well-established datasets. Authors selected tasks that are more diverse and more difficult than SentEval. Otherwise, the design of the benchmark is similar to SentEval. %Overall, they proved to be more popular and their benchmark is widely used.

The aforementioned benchmarks are limited to the English language. Noteworthy attempts at providing multi-language benchmarks include XNLI dataset \citep{conneau2018xnli}, with the MNLI \citep{N18-1101} dataset translated by professional translators into 14 languages. A~similar effort is XQuAD \citep{artetxe2019crosslingual} which is a~translation of the SQuAD dataset \citep{rajpurkar2016squad} into 10 languages. 

None of these efforts includes Polish. Other resources to evaluate the Polish language understanding models are scarce. Recently, \citet{krasnowska2019empirical} prepared their version of the SentEval probing tasks for the Polish language. However, it is more suited for analyzing the sentence embeddings linguistic properties than assessing their quality. 

The PolEval \citep{wawer2017results, kobylinski2017results, ogr:kob:18:poleval, ogr:kob:19:poleval}\footnote{\url{http://poleval.pl}} platform organizes an annual competition in natural language processing for the Polish language. During the first three editions, it assembled 11 diverse tasks and attracted over 40 teams. It could serve as the natural benchmark for the Polish language understanding, but it lacks the common interface for all tasks, making it difficult and time-consuming to use. We include one of the PolEval tasks into the KLEJ Benchmark.

Recently \citet{dadas2019evaluation} introduced a~benchmark similar to the KLEJ benchmark proposed in this paper. It contains two sentiment analysis tasks, topic classification and a~Polish translation of the SICK dataset \citep{marco_marelli_2014_2787612}. Similarly to their work, we use the same sentiment analysis dataset, but transform it into a more difficult task. We also use the analogous dataset to SICK but created from scratch for the Polish language. Finally, we considered the topic classification task to be too easy to include into the proposed benchmark. Overall, KLEJ benchmark consists of nine tasks. They are more diverse, cover a~wider range of objectives and evaluate not only single sentences but also whole paragraphs.

\section{Tasks}
\label{tasks}

KLEJ consists of nine Polish language understanding tasks. Similarly to GLUE, we choose tasks from different domains and with different objectives. In contrast to previous benchmarks, we include several tasks that take multiple sentences as input. We decided to focus on tasks which have relatively small datasets -- most of them have less than 10k training examples. Moreover, some tasks require extensive external knowledge to solve them. Such a~setup promotes knowledge transfer techniques like transfer learning, instead of training separate models for each task from scratch. In effect, KLEJ supports the goal of creating a~general model for the Polish language understanding. We present all tasks in the following sections and summarize them in Table \ref{tab:overview}.

\subsection{NKJP-NER}
We use the human-annotated part of the NKJP (\textit{Narodowy Korpus J\k{e}zyka Polskiego}, eng. \textit{National Corpus of Polish}) \citep{przepiorkowski2012narodowy} to create the named entity classification task. 

The original dataset consists of 85k sentences, randomly selected from a~much larger, balanced and representative corpus of contemporary Polish. We use existing human-annotations of named entities to convert the dataset into a~named entity classification task. First, we filter out all sentences with entities of more than one type. Then, we randomly assigned sentences into training, development and test sets in such a way, that each named entity appears only in one of the three splits. We decided to split the sentences based on named entities to make the task more difficult. To increase class balance, we undersample the \texttt{persName} class and merge \texttt{date} and \texttt{time} classes. Finally, we sample sentences without any named entity to represent the \texttt{noEntity} class.

The final dataset consists of 20k sentences and six classes. The task is to predict the presence and type of a~named entity. Although the named entity classification task differs from traditional NER task, it has a comparable difficulty and evaluates similar aspects of language understanding. At the same time, it follows the common technical interface as other KLEJ tasks, which makes it easy to use. We use accuracy for evaluation.

\subsection{CDSC}
The Compositional Distributional Semantics Corpus \citep{wroblewska2017polish} consists of pairs of sentences which are human-annotated for semantic relatedness and entailment. Although the main design of the dataset is inspired by SICK, it differs in details. As in SICK, the sentences come from image captions, but the set of chosen images is much more diverse as they come from 46 thematic groups. We prepared two KLEJ tasks based on the CDS Corpus.

\subsubsection{CDSC-R}
The first task is to predict relatedness between a pair of sentences, ranging from 0 (not related) to 5 (very related). The score is the average of scores assigned by three human annotators. We use the Spearman correlation to measure the performance of the model.

\subsubsection{CDSC-E}
The second task uses the textual entailment annotations to predict if the premise entails the hypothesis (entailment), negates the hypothesis (contradiction), or is unrelated (neutral). Even though there is an imbalanced label distribution (most of them are neutral) we follow \citet{krasnowska2019empirical} and use accuracy as an evaluation metric.

\subsection{CBD}
The Cyberbullying Detection task \citep{ptaszynski2019results} was a~part of the 2019 edition of the PolEval competition\footnote{\url{http://2019.poleval.pl/index.php/tasks/task6}}. The goal is to predict whether a~given Twitter message is a~case of cyberbullying. We use the dataset as-is and use F1-Score to measure the performance of a~given model, following the original design of the task.

\subsection{PolEmo2.0}
The PolEmo2.0 \citep{kocon-etal-2019-multi} is a~dataset of online consumer reviews from four different domains, namely: medicine, hotels, products and university. It is human-annotated on a level of full reviews, as well as individual sentences. It consists of over 8000 reviews, about 85\% of which are from the medicine and hotel domains. Each review is annotated with one of four labels: positive, negative, neutral or ambiguous. The task is to predict the correct label.

We use the PolEmo2.0 dataset to form two tasks. Both of them use the same training dataset, i.e. reviews from medicine and hotel domains, but are evaluated on a~different test set.

\subsubsection{In-Domain}
In the first task, we use accuracy to evaluate model performance within the in-domain context, i.e. on a~test set of reviews from medicine and hotels domains.

\subsubsection{Out-of-Domain}
In the second task, we test the model on out-of-domain reviews, i.e. from product and university domains. Since the original test sets for those domains are scarce (50 reviews each) we decided to use the original out-of-domain training set of 900 reviews for testing purposes and create the new split of development and test sets. As a result, the task consists of 1000 reviews, which is comparable in size to the in-domain test dataset of 1400 reviews.

\subsection{Czy wiesz?}
The \textit{Czy wiesz?} (eng. Did you know?) dataset \citep{marcinczuk2013open} consists of almost 5k question-answer pairs obtained from \textit{Czy wiesz...} section of Polish Wikipedia. Each question is written by a~Wikipedia collaborator and is answered with a~link to a~relevant Wikipedia article. 

The authors of the dataset used it to build a~Question Answering system. Then, they evaluated the system using 1.5k questions. For each question, they took the top 10 system responses and manually annotated if the answer was correct. Positive responses to 250 questions were further processed and only relevant continuous parts of responses were selected by human annotators. Following this procedure, we have manually extracted shorter responses from the remaining positive examples. Finally, we used these annotations to create positive question-answer pairs.

To select the most difficult negative answers, we used the byte-pair encoding (BPE) \citep{sennrich2016bpe} token overlap between a~question and a~possible answer%\footnote{We also checked other metrics for text similarity: Jaccard index and word-based lexical overlap, but the subword units worked best in revealing the difference between positively and negatively annotated examples.}
. For each question, we took only four most similar negatives and removed ones with a~similarity metric score below the threshold $\tau = 0.3$. On average, the negative answers were much longer than the positive ones. Since it could be potentially exploited by the model, we decided to balance the length of the positive and negative answers. To sample the most relevant part of a~negative example, we used BPE based metric with an additional penalty for the number of sentences: 
$$\widehat{\rm sim}_{\rm BPE} = \frac{\rm sim_{BPE}}{{1.2}^{\rm \#sents}}$$

The task is to predict if the answer to the given question is correct or not. Since the dataset is highly imbalanced, we chose F1-score metric.

\subsection{PSC}
The Polish Summaries Corpus (PSC) \citep{ogro:kop:14:lrec} is a~dataset of summaries for 569 news articles. For each article, the human annotators created five extractive summaries by choosing approximately 5\% of the original text. Each summary was created by a different annotator. The subset of 154 articles was also supplemented with additional five abstractive summaries each, i.e. not created from the fragments of the original article.

Based on PSC we formulate a~text-similarity task. We generate the positive pairs (i.e. referring to the same article) using only those news articles which have both extractive and abstractive summaries. We match each extractive summary with two least similar abstractive ones of the same article. We use the same similarity metric as in the preparation of the \textit{Czy wiesz?} dataset, calculating the BPE token overlap between the extractive and abstractive summary.

To create negative pairs, we follow a~similar procedure. For each extractive summary, we find two most similar abstractive summaries, but from different articles. We remove examples with similarity below the threshold $\tau = 0.15$. To increase the difficulty and diversity of the task, we filter out multiple abstracts from the same article. As a~result, there is at most one negative pair created from each pair of articles.

In total, we obtain around 4k examples. We randomly split the dataset into train and test based on the articles of the extracts to further increase the task's difficulty. For evaluation, we use F1-score.

\subsection{AR}
We introduce a~new sentiment analysis dataset, named Allegro Reviews (AR), extracting 12k product reviews from Allegro.pl - a popular e-commerce marketplace. Each review is at least 50 words long and has a~rating on a~scale from one (negative review) to five (positive review). The task is to predict the rating of a~given review. 

To counter slight class imbalance in the dataset, we propose to evaluate models using $wMAE$, i.e. macro-average of the mean absolute error per class. Additionally, we transform the rating to be between zero and one and report $1 - wMAE$ to ensure consistent metric interpretation between tasks.

\section{Baselines}
\label{baselines}

In this section, we present an overview of several baseline models, which we evaluated using the KLEJ benchmark. We divide these models into three main groups: (1) the LSTM-based \citep{bilstm:1997} models using pretrained word embeddings, (2) models based on Transformer architecture and (3) BERT model trained on Polish corpora. We also include the simple baseline by sampling targets from a training set.

\subsection{LSTM-based models}
We chose the standard Bidirectional LSTM text encoder as the base architecture. Following the GLUE experiments setup \citep{wang2019glue} we trained it jointly as the multi-task learner on all KLEJ tasks.

The architecture consists of two parts: a~shared sentence encoder and a~task specific classifier. The sentence representation model is a~two layer BiLSTM with 1024 hidden units, 300 dimensional word embeddings and max pooling. The classifier is an~MLP with 512 dimensional hidden layer.

We perform training in two stages. First, we pretrain the whole model in a~multi-task scheme. In the second stage, we freeze the sentence encoder and fine-tune the classifiers separately for each task. The initial learning rate in both phases was set to $10^{-4}$ with linear decay down to $10^{-5}$. Pretraining progress is measured by the macro average of all task metrics. We train models with a~batch size of 128, except for the ELMo version, which is trained with a~batch size of 64. For tasks without development set, we use 10\% of training examples as validation data.

We used \texttt{jiant} \citep{wang2019jiant} library to train the LSTM-based models and report the median performance of 5 runs.

\subsubsection{Vanilla BiLSTM}
\label{plainBiLSTM}
The simplest version of the LSTM-based models is a~BiLSTM sentence encoder with an~MLP classifier trained from scratch without any form of transfer learning, i.e. without the usage of pretrained word embeddings.

\subsubsection{fastText}
Before contextual word embeddings became widely adopted, models were enhanced with pretrained word vectors. To evaluate their impact on KLEJ tasks, we initialize word embeddings with fastText \citep{bojanowski2016enriching} trained on Common Crawl and Wikipedia for Polish language \citep{grave2018learning}. %The rest of the architecture and training setup remains the same as for \ref{plainBiLSTM}

\subsubsection{ELMo}
ELMo \citep{Peters:2018} is a~bidirectional language model using character-level convolutions. In contrast to fastText, ELMo's embeddings capture word-level semantics in a~context of the whole sentence.

We conducted more thorough experiments with ELMo embeddings. During the fine-tuning stage in training on a~downstream KLEJ task, we modified the sentence encoder parameters and trained the entire architecture with only a~word embedding's weights unmodified. Additionally, we experimented with the attention mechanism \citep{conneau-etal-2017-supervised} between all words in tasks with a pair of sentences.

We use publicly available pretrained ELMo weights for Polish language \citep{clarin2019elmo}.

\subsection{Transformer-based models}
Recently, the best results on the GLUE benchmark were obtained by Transformer-based models inspired by the Bidirectional Encoder Representations (BERT) model. All of them are pretrained on large text corpora using some variant of Masked Language Model (MLM) objective. In this section, we describe three such models: Multilingual BERT, XLM \citep{lample2019cross} and Slavic-BERT \citep{arkhipov-etal-2019-tuning}. At the time of writing this paper, these are the only available Transformer-based models that were trained with Polish text.

To evaluate these models we fine-tune them on each task separately. For training we used the \texttt{transformers} \citep{Wolf2019HuggingFacesTS} library. All models were trained for 4 epochs with a~batch size of 32 and using a~linearly decaying learning rate scheme starting at $2 \times 10^{-5}$ with a~100 iteration warm-up. We use Adam optimizer with parameters: $\beta_{1} = 0.9$, $\beta_{2} = 0.999$, $\epsilon = 10^{-8}$. We report the median performance of 5 runs.

\subsubsection{Multilingual BERT}
The BERT is a~popular model based on the Transformer architecture trained using MLM and Next Sentence Prediction (NSP) objectives. We use the Multilingual Cased BERT model%\footnote{\url{https://github.com/google-research/bert/blob/master/multilingual.md}}
, which was trained on 104 languages (including Polish), selecting ones with the largest among all Wikipedia corpora. It uses the shared WordPiece \citep{wu2016google} tokenizer with the vocabulary size of 110k. 

\subsubsection{XLM}
The Cross-lingual Language Model (XLM) is based on BERT. It differs from BERT in that it does not use NSP objective, has more layers (16 vs 12), more attention heads (16 vs 12), larger hidden layers size (1280 vs 768) and a~larger vocabulary (200k vs 110k). Moreover, the vocabulary is learned on a~corpus for which the most popular languages were undersampled to balance the number of tokens between high- and low-resource languages. We use the XLM-17 model, which was trained on Wikipedia for 17 languages (including Polish).

\subsubsection{Slavic-BERT}
The Slavic-BERT is a~BERT model trained on four Slavic languages (Polish, Czech, Russian, and Bulgarian). Contrary to previous models, \citet{arkhipov-etal-2019-tuning} used not only Wikipedia but also the Russian News corpus. To avoid costly pretraining, the model was initialized with Multilingual BERT.

\subsection{HerBERT}
None of the above models was optimized for Polish and all of them were trained on Wikipedia only. We decided to combine several publicly available corpora and use them to train a~Transformer-based model specifically for the Polish language.

\subsubsection{Corpora}
In this section, we describe the corpora on which we trained our model. Due to copyright constraints we were not able to use the National Corpus of Polish (NKJP), the most commonly known Polish corpus. Instead, we combined several other publicly available corpora and created a~larger, but less representative corpus.

\begin{table}[!ht]
\renewcommand*{\arraystretch}{1.15}
\setlength{\tabcolsep}{0.38em}
\centering
\begin{tabular}{lrrr}

      \toprule
      \bf{Corpus} & \bf{Tokens} & \bf{Texts} & \bf{Avg len} \\
      \toprule
      NKJP & 1357M & 3.9M & 348 \\
      \midrule
      \midrule
      OSCAR & 6710M & 145M & 46 \\
      Open Subtitles & 1084M & 1.1M & 985 \\
      Wikipedia & 260M & 1.5M & 190 \\
      Wolne Lektury & 41M & 5.5k & 7450 \\
      Allegro Articles & 18M & 33k & 552 \\
      \midrule
      Total & 8113M & 150M & 54 \\
      \bottomrule

\end{tabular}
\caption{Overview of corpora used to train HerBERT compared to the NKJP. \textit{Avg len} is the average number of tokens per document in each corpus.}
\label{tab:corpora}
\end{table}

\paragraph{Wikipedia} Polish version is among the top 10 largest Wikipedia versions. However, it is still relatively small compared to the English one (260M vs 3700M words).
To extract a~clean corpus from the raw Wikipedia dump, we used the tools provided by XLM.\footnote{\url{https://github.com/facebookresearch/XLM}} However, we did not lowercase the text and did not remove diacritics.

\paragraph{Wolne Lektury}(eng. \emph{Free Readings}) is an online repository of over 5k books, written by Polish authors or translated into Polish. %Documents are available under the principles set out in the Definition of Free Cultural Property\footnote{https://freedomdefined.org/Definition}. 
Although the majority of the books in the dataset were written in the 19th or 20th century and they might not be fully representative of the contemporary Polish, they are free to download and can be used as a~text corpus.

\paragraph{Open Subtitles} is a~multilingual parallel corpus based on movie and TV subtitles \citep{LISON16.947} from the \texttt{opensubtitles.org} website. As a~result, it contains very specific, mostly conversational text consisting of short sentences. Since the translations are community-sourced, they may be of substandard quality. The Polish part of the dataset is relatively large compared to the other corpora (see Table \ref{tab:corpora}).

\paragraph{OSCAR} is a~multilingual corpus created by \citet{ortizsuarez:hal-02148693} based on Common Crawl\footnote{\url{http://commoncrawl.org/}}. The original dataset lacks information about the language used in particular documents. Categorization to specific languages was automated by a~classifier, splitting whole Common Crawl into many monolingual corpora. Duplicates were removed from the dataset to increase its quality. We only use the Polish part of the corpus and use texts longer than 100 words.

\paragraph{Allegro Articles} Additionally, we obtained over 30k articles from Allegro.pl - a popular e-commerce marketplace. They contain product reviews, shopping guides and other texts from the e-commerce domain. It is the smallest corpus we've used, but it contains high-quality documents from the domain of our interest.

\begin{table*}
\renewcommand*{\arraystretch}{1.15}
\centering
\begin{tabular}{l|c|ccccccccc}

    \toprule
    \bf{Model} & \rot{\bf{AVG }} & \rot{\bf{NKJP-NER }} & \rot{\bf{CDSC-E }} & \rot{\bf{CDSC-R }} & \rot{\bf{CBD }} & \rot{\bf{PolEmo2.0-IN }} & \rot{\bf{PolEmo2.0-OUT }} & \rot{\bf{Czy wiesz? }} & \rot{\bf{PSC }} & \rot{\bf{AR }} \\
    \toprule
    % \multicolumn{11}{c}{Simple Baseline} \\
    % \midrule
    Random & 28.3 & 20.7 & 59.2 & 0.9 & 11.2 & 27.8 & 28.5 & 18.9 & 30.4 & 56.9 \\
    \midrule
    % \multicolumn{11}{c}{LSTM-based Models} \\
    % \midrule
    LSTM & 63.0 & 45.0 & 87.5 & 84.7 & 20.7 & 79.6 & 60.7 & 22.3 & 84.4 & 81.8 \\
    LSTM + fastText & 67.7 & 67.3 & 87.8 & 81.6 & 32.8 & 83.2 & 61.1 & 27.5 & 86.5 & 81.4 \\
    LSTM + ELMo & 76.6 & 93.0 & 88.9 & 90.4 & 50.2 & 88.5 & 72.1 & 28.8 & 92.7 & 85.1 \\
    LSTM + ELMo + fine-tune & 76.7 & \bf{93.4} & 89.3 & 91.1 & 47.9 & 87.4 & 70.6 & 30.9 & 93.7 & \bf{86.2} \\
    LSTM + ELMo + attention & 75.8 & 93.0 & 90.0 & 90.3 & 46.8 & 88.8 & 70.2 & 26.0 & 92.1 & 85.4 \\
    \midrule
    % \multicolumn{11}{c}{Transformer-based Models} \\
    % \midrule
    Multi-BERT & 79.5 & 91.4 & \bf{93.8} & 92.9 & 40.0 & 85.0 & 66.6 & \bf{64.2} & 97.9 & 83.3 \\
    Slavic-BERT & 79.8 & 93.3 & 93.7 & \bf{93.3} & 43.1 & 87.1 & 67.6 & 57.4 & \bf{98.3} & 84.3 \\
    XLM-17 & 80.2 & 91.9 & 93.7 & 92.0 & 44.8 & 86.3 & 70.6 & 61.8 & 96.3 & 84.5 \\
    % \midrule
    % \multicolumn{11}{c}{HerBERT} \\
    \midrule
    % HerBERT 60k & 78.9 & 92.0 & 92.9 & 90.9 & 47.4 & 88.1 & 73.3 & 46.6 & 93.7 & 85.0 \\
    % HerBERT 120k & 79.6 & 92.2 & 92.9 & 91.7 & 48.9 & 88.4 & 74.5 & 48.3 & 94.4 & 85.2 \\
    HerBERT & \bf{80.5} & 92.7 & 92.5 & 91.9 & \bf{50.3} & \bf{89.2} & \bf{76.3} & 52.1 & 95.3 & 84.5 \\
    \bottomrule

\end{tabular}
\caption{Baseline evaluation on KLEJ benchmark. \textit{AVG} is the average score across all tasks.}
\label{tab:results}
\end{table*}

\subsubsection{Model}
\paragraph{Architecture} HerBERT is a~multi-layer bidirectional Transformer. We use BERT\textsubscript{BASE} architecture configuration with 12 layers, 12 attention heads and hidden dimension of 768.

\paragraph{Loss} We train HerBERT with a~MLM objective. According to the updated version of BERT, we always mask all tokens corresponding to the randomly picked word. Whole word masking objective is more difficult to learn than predicting subword tokens \citep{joshi2019spanbert, louis2019camembert}.

In the original BERT training setup tokens are masked statically during the text preprocessing phase. In HerBERT, we chose to use dynamic token masking, which follows the training setup of the RoBERTa model \citep{liu2019roberta}.

We decided not to use the NSP objective. Previous studies by \citet{yang2019xlnet} and \citet{liu2019roberta} showed that this objective is too easy and does not improve performance on downstream tasks.

\paragraph{Data preprocessing}
We tokenize corpus data into subword tokens using BPE. We learn BPE splits on Wolne Lektury and a~publicly available subset of National Corpus of Polish. We choose these two datasets because of their higher quality compared to the rest of our corpus. We limit the vocabulary size to 50k tokens.

Our datasets contain a~lot of small fragments of coherent text that should be treated as separate documents. We remove degenerated documents that consist of less than 20 tokens from available corpora. Maximal segment length is 512 as it was originally proposed in BERT. We do not accumulate short examples into full 512 token segments because such sequences would be incoherent with frequent topic changes. The only exception to this rule is the Open Subtitles dataset, where subsequent parts of dialogues were connected to form larger documents. The aforementioned training setup gives us a~slightly better performance on downstream tasks than simply selecting all available data.

\paragraph{Hyperparameters}
We train HerBERT using Adam optimizer \citep{Kingma2015Adam} with parameters: $\beta_{1} = 0.9$, $\beta_{2} = 0.999$, $\epsilon = 10^{-8}$. We use learning rate burn-in over the first $500$ steps, reaching a peak value of $10^{-4}$; the learning rate is then linearly decayed for the rest of the training. We train the model with a~batch size of $570$. HerBERT was trained for $180$k steps, without showing signs of saturation.

\section{Evaluation}
\label{evaluation}

We first compare models based on their average performance. Even though it is not the definite metric to compare models, especially as not all tasks are equally difficult, it gives a general notion of the model performance across all tasks.

In comparison to baselines based on the LSTM architecture, Transformer-based models clearly show superior performance. The only exception is ELMo, which achieves competitive results on many tasks. On two of them, the fine-tuned ELMo model achieves the best score. In general, the evaluation shows major shortcomings of multilingual pretrained BERT models for Polish, and possibly other low resource languages. Overall, every LSTM-based baseline is still on average worse than any of the tested Transformer-based models. 

%Attention mechanism improved results on several tasks with pair of texts but in general it's effect is negligible. Different from our observations, Wang \citep{wang2019glue} shows that ELMo baselines with attention achieve best performance on GLUE tasks.  

The KLEJ benchmark was designed to require additional knowledge to promote general language understanding models. As expected, we observe significant increases of models quality when using pretrained word embeddings. The vanilla LSTM model achieves the average score of 63.0 while supplying it with the fastText embeddings boosts performance to 67.7. Usage of more recent, contextualized embeddings (ELMo) increases the score to 76.6. 

Focusing on fewer languages seems to result in a~better model. The Slavic-BERT has higher scores than Multi-BERT on seven out of nine tasks. However, without a detailed ablation study, it is difficult to infer the main reason resulting in better performance. It can also be attributed to a better tokenizer, additional Russian News corpus or longer training (the Slavic-BERT was initialized with Multi-BERT weights).

The training corpus seems to play an important role in the performance of a~downstream task. Both HerBERT and ELMo models were trained mainly on web crawled texts and they excel at tasks from an online domain (\textit{CBD}, \textit{PolEmo-IN}, \textit{PolEmo-OUT}, and \textit{AR}). On the other hand, the other Transformer-based models are superior on the \textit{Czy wiesz?} task. It can be related to the fact that it is a~Wikipedia-based question-answering task and the aforementioned models were trained mainly on Wikipedia corpus. Interestingly, the Slavic-BERT, which was additionally trained on Russian News corpus, has a lower score on the \textit{Czy wiesz?} task than Multi-BERT and XLM-17.

HerBERT achieves highly competitive results compared to the other Transformer-based models. It has the best performance on average and achieves state-of-the-art results on three tasks, \textit{PolEmo-IN}, \textit{PolEmo-OUT} and \textit{CBD}. Moreover, HerBERT has the smallest performance gap between \textit{PolEmo-IN} and \textit{PolEmo-OUT}, which suggests better generalization across domains. Compared to the other Transformer-based models it performs poorly on \textit{Czy wiesz?} and \textit{PSC} tasks.

The KLEJ benchmark proved to be challenging and diverse. There is no clear winner among evaluated models; different models perform better at different tasks. It suggests that the KLEJ benchmark is far from being solved, and it can be used to evaluate and compare future models.

\section{Conclusion}
\label{conclusion}
% We introduce the KLEJ benchmark, a~comprehensive evaluation tool for the Polish language understanding. It proves to be both challenging and diverse, as there is no single model that performs best on all tasks. We also present HerBERT, a~Transformer-based model trained specifically for Polish and compare it with other LSTM and Transformer-based models. We find that it is best on average and achieves highest scores on three tasks. We plan to continue the work on HerBERT and use the KLEJ benchmark to guide its development.

We introduce the KLEJ benchmark, a~comprehensive set of evaluation tasks for the Polish language understanding. Its goal is to drive the development of better NLU models, so careful selection of tasks was crucial. We mainly focused on a variety of text genres, objectives, text lengths, and difficulties, which allows us to assess the models across different axes. As a result, KLEJ benchmark proves to be both challenging and diverse, as there is no single model that outperforms others on all tasks.

We find it equally important to provide a common evaluation interface for all the tasks. For that purpose, many existing resources had to be adapted, either automatically (\textit{NKJP-NER}, \textit{PSC}) or manually (\textit{Czy wiesz?}), to make it easier to use.

It's worth mentioning that the main weakness of creating such benchmarks is focusing only on the model performance and not the model efficiency, e.g. in terms of training data, speed or a number of parameters. It seems reasonable to derive additional benchmarks by requiring a given level of efficiency from participating models. We leave it as future work.

We also present HerBERT, a~Transformer-based model trained specifically for Polish and compare it with other LSTM- and Transformer-based models. We find that it is the best on average and achieves highest scores on three tasks. We plan to continue the work on HerBERT and use the KLEJ benchmark to guide its development.

\bibliography{acl2020}

\begin{thebibliography}{42}
\expandafter\ifx\csname natexlab\endcsname\relax\def\natexlab#1{#1}\fi

\bibitem[{Arkhipov et~al.(2019)Arkhipov, Trofimova, Kuratov, and
  Sorokin}]{arkhipov-etal-2019-tuning}
Mikhail Arkhipov, Maria Trofimova, Yuri Kuratov, and Alexey Sorokin. 2019.
\newblock \href {https://doi.org/10.18653/v1/W19-3712} {Tuning multilingual
  transformers for language-specific named entity recognition}.
\newblock In \emph{Proceedings of the 7th Workshop on Balto-Slavic Natural
  Language Processing}, pages 89--93, Florence, Italy. Association for
  Computational Linguistics.

\bibitem[{Artetxe et~al.(2019)Artetxe, Ruder, and
  Yogatama}]{artetxe2019crosslingual}
Mikel Artetxe, Sebastian Ruder, and Dani Yogatama. 2019.
\newblock \href {http://arxiv.org/abs/1910.11856} {On the cross-lingual
  transferability of monolingual representations}.

\bibitem[{Bojanowski et~al.(2016)Bojanowski, Grave, Joulin, and
  Mikolov}]{bojanowski2016enriching}
Piotr Bojanowski, Edouard Grave, Armand Joulin, and Tomas Mikolov. 2016.
\newblock Enriching word vectors with subword information.
\newblock \emph{arXiv preprint arXiv:1607.04606}.

\bibitem[{Conneau and Kiela(2018)}]{conneau2018senteval}
Alexis Conneau and Douwe Kiela. 2018.
\newblock Senteval: An evaluation toolkit for universal sentence
  representations.
\newblock In \emph{Proceedings of the Eleventh International Conference on
  Language Resources and Evaluation (LREC-2018)}.

\bibitem[{Conneau et~al.(2017)Conneau, Kiela, Schwenk, Barrault, and
  Bordes}]{conneau-etal-2017-supervised}
Alexis Conneau, Douwe Kiela, Holger Schwenk, Lo{\"\i}c Barrault, and Antoine
  Bordes. 2017.
\newblock \href {https://doi.org/10.18653/v1/D17-1070} {Supervised learning of
  universal sentence representations from natural language inference data}.
\newblock In \emph{Proceedings of the 2017 Conference on Empirical Methods in
  Natural Language Processing}, pages 670--680, Copenhagen, Denmark.
  Association for Computational Linguistics.

\bibitem[{Conneau et~al.(2018)Conneau, Rinott, Lample, Williams, Bowman,
  Schwenk, and Stoyanov}]{conneau2018xnli}
Alexis Conneau, Ruty Rinott, Guillaume Lample, Adina Williams, Samuel~R.
  Bowman, Holger Schwenk, and Veselin Stoyanov. 2018.
\newblock Xnli: Evaluating cross-lingual sentence representations.
\newblock In \emph{Proceedings of the 2018 Conference on Empirical Methods in
  Natural Language Processing}. Association for Computational Linguistics.

\bibitem[{Dadas et~al.(2019)Dadas, Pere\l{}kiewicz, and
  Po\'{s}wiata}]{dadas2019evaluation}
S\l{}awomir Dadas, Micha\l{} Pere\l{}kiewicz, and Rafa\l{} Po\'{s}wiata. 2019.
\newblock \href {http://arxiv.org/abs/1910.11834} {Evaluation of sentence
  representations in polish}.

\bibitem[{Devlin et~al.(2019)Devlin, Chang, Lee, and
  Toutanova}]{devlin2019bert}
Jacob Devlin, Ming-Wei Chang, Kenton Lee, and Kristina Toutanova. 2019.
\newblock Bert: Pre-training of deep bidirectional transformers for language
  understanding.
\newblock In \emph{Proceedings of the 2019 Conference of the North American
  Chapter of the Association for Computational Linguistics: Human Language
  Technologies, Volume 1 (Long and Short Papers)}, pages 4171--4186.

\bibitem[{Grave et~al.(2018)Grave, Bojanowski, Gupta, Joulin, and
  Mikolov}]{grave2018learning}
Edouard Grave, Piotr Bojanowski, Prakhar Gupta, Armand Joulin, and Tomas
  Mikolov. 2018.
\newblock Learning word vectors for 157 languages.
\newblock In \emph{Proceedings of the International Conference on Language
  Resources and Evaluation (LREC 2018)}.

\bibitem[{Hochreiter and Schmidhuber(1997)}]{bilstm:1997}
S.~Hochreiter and J.~Schmidhuber. 1997.
\newblock {Long short-term memory}.
\newblock \emph{Neural Computation}, 9(8):1735--1780.

\bibitem[{Howard and Ruder(2018)}]{howard2018universal}
Jeremy Howard and Sebastian Ruder. 2018.
\newblock Universal language model fine-tuning for text classification.
\newblock In \emph{Proceedings of the 56th Annual Meeting of the Association
  for Computational Linguistics (Volume 1: Long Papers)}, pages 328--339.

\bibitem[{Janz(2019)}]{clarin2019elmo}
Arkadiusz Janz. 2019.
\newblock \href {http://hdl.handle.net/11321/690} {{ELMo} embeddings for
  polish}.
\newblock {CLARIN}-{PL} digital repository.

\bibitem[{Joshi et~al.(2019)Joshi, Chen, Liu, Weld, Zettlemoyer, and
  Levy}]{joshi2019spanbert}
Mandar Joshi, Danqi Chen, Yinhan Liu, Daniel~S. Weld, Luke Zettlemoyer, and
  Omer Levy. 2019.
\newblock Spanbert: Improving pre-training by representing and predicting
  spans.
\newblock \emph{arXiv preprint arXiv:1907.10529}.

\bibitem[{Kingma and Ba(2014)}]{Kingma2015Adam}
Diederik~P. Kingma and Jimmy Ba. 2014.
\newblock \href {http://arxiv.org/abs/1412.6980} {Adam: A method for stochastic
  optimization}.
\newblock Cite arxiv:1412.6980Comment: Published as a conference paper at the
  3rd International Conference for Learning Representations, San Diego, 2015.

\bibitem[{Kobyli{\'n}ski and Ogrodniczuk(2017)}]{kobylinski2017results}
{\L}ukasz Kobyli{\'n}ski and Maciej Ogrodniczuk. 2017.
\newblock Results of the poleval 2017 competition: Part-of-speech tagging
  shared task.
\newblock In \emph{Proceedings of the 8th Language and Technology Conference:
  Human Language Technologies as a Challenge for Computer Science and
  Linguistics}, pages 362--366.

\bibitem[{Koco{\'n} et~al.(2019)Koco{\'n}, Mi{\l}kowski, and
  Za{\'s}ko-Zieli{\'n}ska}]{kocon-etal-2019-multi}
Jan Koco{\'n}, Piotr Mi{\l}kowski, and Monika Za{\'s}ko-Zieli{\'n}ska. 2019.
\newblock \href {https://doi.org/10.18653/v1/K19-1092} {Multi-level sentiment
  analysis of {P}ol{E}mo 2.0: Extended corpus of multi-domain consumer
  reviews}.
\newblock In \emph{Proceedings of the 23rd Conference on Computational Natural
  Language Learning (CoNLL)}, pages 980--991, Hong Kong, China. Association for
  Computational Linguistics.

\bibitem[{Krasnowska-Kiera{\'s} and
  Wr{\'o}blewska(2019)}]{krasnowska2019empirical}
Katarzyna Krasnowska-Kiera{\'s} and Alina Wr{\'o}blewska. 2019.
\newblock Empirical linguistic study of sentence embeddings.
\newblock In \emph{Proceedings of the 57th Annual Meeting of the Association
  for Computational Linguistics}, pages 5729--5739.

\bibitem[{Lample and Conneau(2019)}]{lample2019cross}
Guillaume Lample and Alexis Conneau. 2019.
\newblock Cross-lingual language model pretraining.
\newblock \emph{arXiv preprint arXiv:1901.07291}.

\bibitem[{Lison and Tiedemann(2016)}]{LISON16.947}
Pierre Lison and J{\"o}rg Tiedemann. 2016.
\newblock Opensubtitles2016: Extracting large parallel corpora from movie and
  tv subtitles.
\newblock In \emph{Proceedings of the Tenth International Conference on
  Language Resources and Evaluation (LREC 2016)}, Paris, France. European
  Language Resources Association (ELRA).

\bibitem[{Liu et~al.(2019)Liu, Ott, Goyal, Du, Joshi, Chen, Levy, Lewis,
  Zettlemoyer, and Stoyanov}]{liu2019roberta}
Yinhan Liu, Myle Ott, Naman Goyal, Jingfei Du, Mandar Joshi, Danqi Chen, Omer
  Levy, Mike Lewis, Luke Zettlemoyer, and Veselin Stoyanov. 2019.
\newblock Roberta: A robustly optimized bert pretraining approach.
\newblock \emph{arXiv preprint arXiv:1907.11692}.

\bibitem[{Marcinczuk et~al.(2013)Marcinczuk, Ptak, Radziszewski, and
  Piasecki}]{marcinczuk2013open}
Micha{\l} Marcinczuk, Marcin Ptak, Adam Radziszewski, and Maciej Piasecki.
  2013.
\newblock Open dataset for development of polish question answering systems.
\newblock In \emph{Proceedings of the 6th Language \& Technology Conference:
  Human Language Technologies as a Challenge for Computer Science and
  Linguistics, Wydawnictwo Poznanskie, Fundacja Uniwersytetu im. Adama
  Mickiewicza}.

\bibitem[{Marelli et~al.(2014)Marelli, Menini, Baroni, Bentivogli, Bernardi,
  and Zamparelli}]{marco_marelli_2014_2787612}
Marco Marelli, Stefano Menini, Marco Baroni, Luisa Bentivogli, Raffaella
  Bernardi, and Roberto Zamparelli. 2014.
\newblock \href {https://doi.org/10.5281/zenodo.2787612} {The sick (sentences
  involving compositional knowledge) dataset for relatedness and entailment}.

\bibitem[{{Martin} et~al.(2019){Martin}, {Muller}, {Ortiz Su{\'a}rez},
  {Dupont}, {Romary}, {Villemonte de la Clergerie}, {Seddah}, and
  {Sagot}}]{louis2019camembert}
Louis {Martin}, Benjamin {Muller}, Pedro~Javier {Ortiz Su{\'a}rez}, Yoann
  {Dupont}, Laurent {Romary}, {\'E}ric {Villemonte de la Clergerie}, Djam{\'e}
  {Seddah}, and Beno{\^\i}t {Sagot}. 2019.
\newblock \href {http://arxiv.org/abs/1911.03894} {{CamemBERT: a Tasty French
  Language Model}}.
\newblock \emph{arXiv e-prints}, page arXiv:1911.03894.

\bibitem[{McCann et~al.(2018)McCann, Keskar, Xiong, and
  Socher}]{McCann2018decaNLP}
Bryan McCann, Nitish~Shirish Keskar, Caiming Xiong, and Richard Socher. 2018.
\newblock The natural language decathlon: Multitask learning as question
  answering.
\newblock \emph{arXiv preprint arXiv:1806.08730}.

\bibitem[{Ogrodniczuk and Kobyli{\'n}ski(2018)}]{ogr:kob:18:poleval}
Maciej Ogrodniczuk and {\L}ukasz Kobyli{\'n}ski, editors. 2018.
\newblock \href {http://poleval.pl/files/poleval2018.pdf} {\emph{{Proceedings
  of the PolEval 2018 Workshop}}}. Institute of Computer Science, Polish
  Academy of Sciences, Warsaw, Poland.

\bibitem[{Ogrodniczuk and Kobyli{\'n}ski(2019)}]{ogr:kob:19:poleval}
Maciej Ogrodniczuk and {\L}ukasz Kobyli{\'n}ski, editors. 2019.
\newblock \href {http://2019.poleval.pl/files/poleval2019.pdf}
  {\emph{{Proceedings of the PolEval 2019 Workshop}}}. Institute of Computer
  Science, Polish Academy of Sciences, Warsaw, Poland.

\bibitem[{Ogrodniczuk and Kope{\'c}(2014)}]{ogro:kop:14:lrec}
Maciej Ogrodniczuk and Mateusz Kope{\'c}. 2014.
\newblock The {P}olish {S}ummaries {C}orpus.
\newblock In \emph{Proceedings of the Ninth International {C}onference on
  {L}anguage {R}esources and {E}valuation, {LREC}~2014}.

\bibitem[{Ortiz~Su{\'a}rez et~al.(2019)Ortiz~Su{\'a}rez, Sagot, and
  Romary}]{ortizsuarez:hal-02148693}
Pedro~Javier Ortiz~Su{\'a}rez, Beno{\^i}t Sagot, and Laurent Romary. 2019.
\newblock \href {https://hal.inria.fr/hal-02148693} {{Asynchronous Pipeline for
  Processing Huge Corpora on Medium to Low Resource Infrastructures}}.
\newblock In \emph{{7th Workshop on the Challenges in the Management of Large
  Corpora (CMLC-7)}}, Cardiff, United Kingdom. {Leibniz-Institut f{\"u}r
  Deutsche Sprache}.

\bibitem[{Peters et~al.(2018)Peters, Neumann, Iyyer, Gardner, Clark, Lee, and
  Zettlemoyer}]{Peters:2018}
Matthew~E. Peters, Mark Neumann, Mohit Iyyer, Matt Gardner, Christopher Clark,
  Kenton Lee, and Luke Zettlemoyer. 2018.
\newblock Deep contextualized word representations.
\newblock In \emph{Proc. of NAACL}.

\bibitem[{Przepi{\'o}rkowski(2012)}]{przepiorkowski2012narodowy}
Adam Przepi{\'o}rkowski. 2012.
\newblock \emph{Narodowy korpus j{\k{e}}zyka polskiego}.
\newblock Naukowe PWN.

\bibitem[{Ptaszynski et~al.(2019)Ptaszynski, Pieciukiewicz, and
  Dyba{\l}a}]{ptaszynski2019results}
Michal Ptaszynski, Agata Pieciukiewicz, and Pawe{\l} Dyba{\l}a. 2019.
\newblock Results of the poleval 2019 shared task 6: First dataset and open
  shared task for automatic cyberbullying detection in polish twitter.
\newblock \emph{Proceedings of the PolEval 2019 Workshop}, page~89.

\bibitem[{Rajpurkar et~al.(2016)Rajpurkar, Zhang, Lopyrev, and
  Liang}]{rajpurkar2016squad}
Pranav Rajpurkar, Jian Zhang, Konstantin Lopyrev, and Percy Liang. 2016.
\newblock Squad: 100,000+ questions for machine comprehension of text.
\newblock In \emph{Proceedings of the 2016 Conference on Empirical Methods in
  Natural Language Processing}, pages 2383--2392.

\bibitem[{Rico~Sennrich and Birch.(2016)}]{sennrich2016bpe}
Barry~Haddow Rico~Sennrich and Alexandra Birch. 2016.
\newblock Neural machine translation of rare words with subword units.
\newblock \emph{In Association for Computational Linguistics (ACL)}, pages
  1715--1725.

\bibitem[{Vaswani et~al.(2017)Vaswani, Shazeer, Parmar, Uszkoreit, Jones,
  Gomez, Kaiser, and Polosukhin}]{vaswani2017attention}
Ashish Vaswani, Noam Shazeer, Niki Parmar, Jakob Uszkoreit, Llion Jones,
  Aidan~N Gomez, \L~ukasz Kaiser, and Illia Polosukhin. 2017.
\newblock \href
  {http://papers.nips.cc/paper/7181-attention-is-all-you-need.pdf} {Attention
  is all you need}.
\newblock In I.~Guyon, U.~V. Luxburg, S.~Bengio, H.~Wallach, R.~Fergus,
  S.~Vishwanathan, and R.~Garnett, editors, \emph{Advances in Neural
  Information Processing Systems 30}, pages 5998--6008. Curran Associates, Inc.

\bibitem[{Wang et~al.(2019{\natexlab{a}})Wang, Singh, Michael, Hill, Levy, and
  Bowman}]{wang2019glue}
Alex Wang, Amanpreet Singh, Julian Michael, Felix Hill, Omer Levy, and
  Samuel~R. Bowman. 2019{\natexlab{a}}.
\newblock {GLUE}: A multi-task benchmark and analysis platform for natural
  language understanding.
\newblock In the Proceedings of ICLR.

\bibitem[{Wang et~al.(2019{\natexlab{b}})Wang, Tenney, Pruksachatkun, Yu, Hula,
  Xia, Pappagari, Jin, McCoy, Patel, Huang, Phang, Grave, Liu, Kim, Htut,
  F'{e}vry, Chen, Nangia, Mohananey, Kann, Bordia, Patry, Benton, Pavlick, and
  Bowman}]{wang2019jiant}
Alex Wang, Ian~F. Tenney, Yada Pruksachatkun, Katherin Yu, Jan Hula, Patrick
  Xia, Raghu Pappagari, Shuning Jin, R.~Thomas McCoy, Roma Patel, Yinghui
  Huang, Jason Phang, Edouard Grave, Haokun Liu, Najoung Kim, Phu~Mon Htut,
  Thibault F'{e}vry, Berlin Chen, Nikita Nangia, Anhad Mohananey, Katharina
  Kann, Shikha Bordia, Nicolas Patry, David Benton, Ellie Pavlick, and
  Samuel~R. Bowman. 2019{\natexlab{b}}.
\newblock \texttt{jiant} 1.2: A software toolkit for research on
  general-purpose text understanding models.
\newblock \url{http://jiant.info/}.

\bibitem[{Wawer and Ogrodniczuk(2017)}]{wawer2017results}
Aleksander Wawer and Maciej Ogrodniczuk. 2017.
\newblock Results of the poleval 2017 competition: Sentiment analysis shared
  task.
\newblock In \emph{8th Language and Technology Conference: Human Language
  Technologies as a Challenge for Computer Science and Linguistics}.

\bibitem[{Williams et~al.(2018)Williams, Nangia, and Bowman}]{N18-1101}
Adina Williams, Nikita Nangia, and Samuel Bowman. 2018.
\newblock \href {http://aclweb.org/anthology/N18-1101} {A broad-coverage
  challenge corpus for sentence understanding through inference}.
\newblock In \emph{Proceedings of the 2018 Conference of the North American
  Chapter of the Association for Computational Linguistics: Human Language
  Technologies, Volume 1 (Long Papers)}, pages 1112--1122. Association for
  Computational Linguistics.

\bibitem[{Wolf et~al.(2019)Wolf, Debut, Sanh, Chaumond, Delangue, Moi, Cistac,
  Rault, Louf, Funtowicz, and Brew}]{Wolf2019HuggingFacesTS}
Thomas Wolf, Lysandre Debut, Victor Sanh, Julien Chaumond, Clement Delangue,
  Anthony Moi, Pierric Cistac, Tim Rault, R'emi Louf, Morgan Funtowicz, and
  Jamie Brew. 2019.
\newblock Huggingface's transformers: State-of-the-art natural language
  processing.
\newblock \emph{ArXiv}, abs/1910.03771.

\bibitem[{Wr{\'o}blewska and
  Krasnowska-Kiera{\'s}(2017)}]{wroblewska2017polish}
Alina Wr{\'o}blewska and Katarzyna Krasnowska-Kiera{\'s}. 2017.
\newblock Polish evaluation dataset for compositional distributional semantics
  models.
\newblock In \emph{Proceedings of the 55th Annual Meeting of the Association
  for Computational Linguistics (Volume 1: Long Papers)}, pages 784--792.

\bibitem[{Wu et~al.(2016)Wu, Schuster, Chen, Le, Norouzi, Macherey, Krikun,
  Cao, Gao, Macherey et~al.}]{wu2016google}
Yonghui Wu, Mike Schuster, Zhifeng Chen, Quoc~V Le, Mohammad Norouzi, Wolfgang
  Macherey, Maxim Krikun, Yuan Cao, Qin Gao, Klaus Macherey, et~al. 2016.
\newblock Google's neural machine translation system: Bridging the gap between
  human and machine translation.
\newblock \emph{arXiv preprint arXiv:1609.08144}.

\bibitem[{Yang et~al.(2019)Yang, Dai, Yang, Carbonell, Salakhutdinov, and
  Le}]{yang2019xlnet}
Zhilin Yang, Zihang Dai, Yiming Yang, Jaime Carbonell, Ruslan Salakhutdinov,
  and Quoc~V Le. 2019.
\newblock Xlnet: Generalized autoregressive pretraining for language
  understanding.
\newblock \emph{arXiv preprint arXiv:1906.0823}.

\end{thebibliography}
\bibliographystyle{acl_natbib}

\end{document}